\documentclass[letterpaper]{article} % DO NOT CHANGE THIS
\usepackage{aaai24}  % DO NOT CHANGE THIS
\usepackage{times}  % DO NOT CHANGE THIS
\usepackage{helvet}  % DO NOT CHANGE THIS
\usepackage{courier}  % DO NOT CHANGE THIS
\usepackage[hyphens]{url}  % DO NOT CHANGE THIS
\usepackage{graphicx} % DO NOT CHANGE THIS
\usepackage{natbib}  % DO NOT CHANGE THIS AND DO NOT ADD ANY OPTIONS TO IT
\usepackage{caption} % DO NOT CHANGE THIS AND DO NOT ADD ANY OPTIONS TO IT
\usepackage{graphicx}
\graphicspath{ {./} }
\usepackage{booktabs} % For better looking tables
\usepackage{multirow} % For multirow feature
\usepackage{xcolor}
\usepackage{enumitem}
\usepackage{algorithm}
\usepackage{algorithmic}

\usepackage{newfloat}
\usepackage{listings}
\DeclareCaptionStyle{ruled}{labelfont=normalfont,labelsep=colon,strut=off} % DO NOT CHANGE THIS
\floatstyle{ruled}
\newfloat{listing}{tb}{lst}{}
\floatname{listing}{Listing}
\title{Evaluating the Capabilities of LLMs for Supporting Anticipatory Impact Assessment}
\author{
    Mowafak Allaham\textsuperscript{\rm 1}, Nicholas Diakopoulos\textsuperscript{\rm 1}\\
}
\affiliations{
    \textsuperscript{\rm 1}Northwestern University\\
    mowafakallaham2021@u.northwestern.edu, nad@northwestern.edu
}

\usepackage{bibentry}
\begin{document}

\maketitle

\begin{abstract}
Gaining insight into the potential negative impacts of emerging Artificial Intelligence (AI) technologies in society is a challenge for implementing anticipatory governance approaches. One approach to produce such insight is to use Large Language Models (LLMs) to support and guide experts in the process of ideating and exploring the range of undesirable consequences of emerging technologies. However, performance evaluations of LLMs for such tasks are still needed, including examining the general quality of generated impacts but also the range of types of impacts produced and resulting biases. In this paper, we demonstrate the potential for generating high-quality and diverse impacts of AI in society by fine-tuning completion models (GPT-3 and Mistral-7B) on a diverse sample of articles from news media and comparing those outputs to the impacts generated by instruction-based (GPT-4 and Mistral-7B-Instruct) models. We examine the generated impacts for coherence, structure, relevance, and plausibility and find that the generated impacts using Mistral-7B, a small open-source model fine-tuned on impacts from the news media, tend to be qualitatively on par with impacts generated using a more capable and larger scale model such as GPT-4. Moreover, we find that impacts produced by instruction-based models had gaps in the production of certain categories of impacts in comparison to fine-tuned models. This research highlights a potential bias in the range of impacts generated by state-of-the-art LLMs and the potential of aligning smaller LLMs on news media as a scalable alternative to generate high quality and more diverse impacts in support of anticipatory governance approaches.
\end{abstract}

\section{Introduction}
\begin{figure*}
\includegraphics[width=\textwidth]{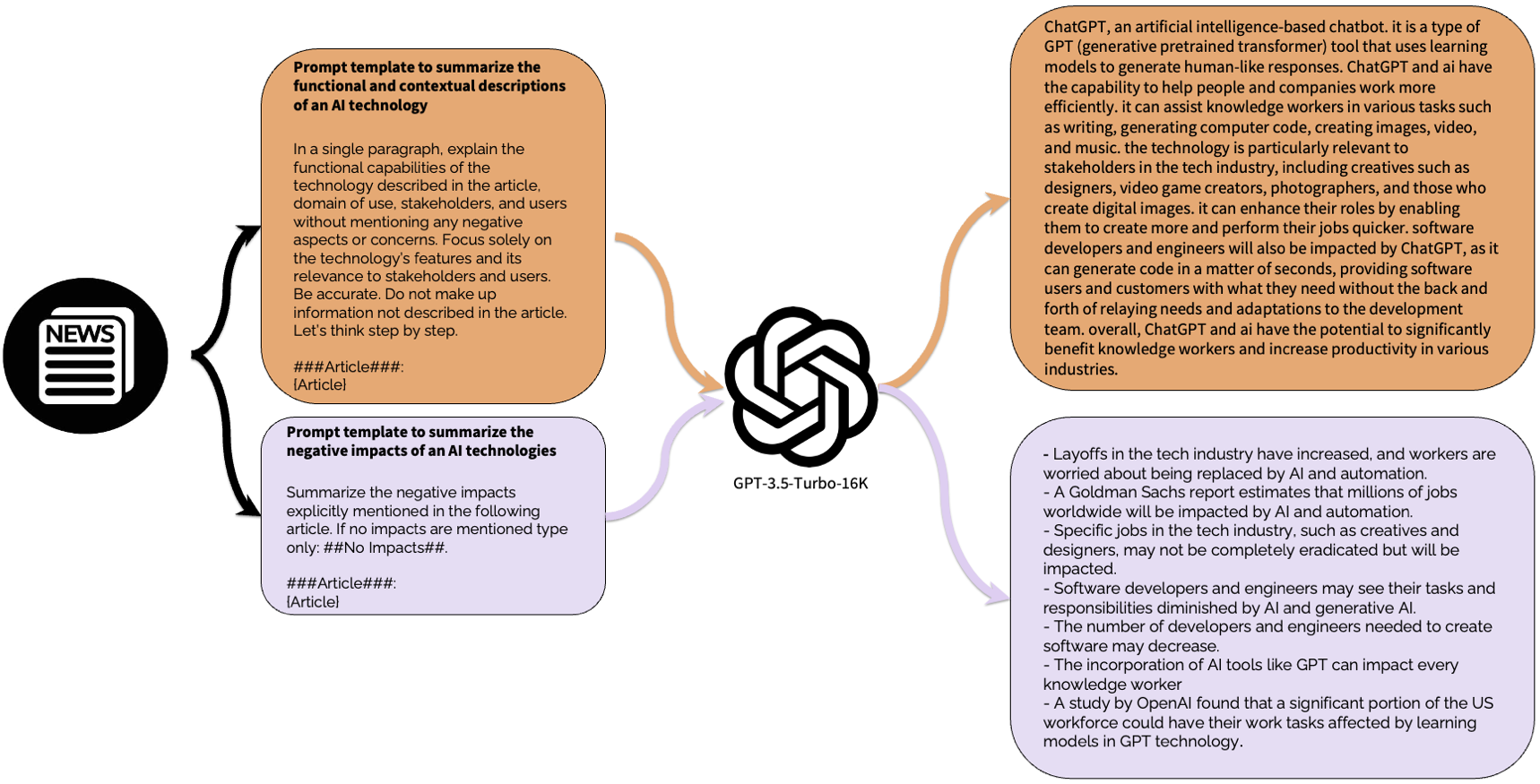}
\caption{An illustration of the workflow to generate the functional and contextual descriptions, as well as the negative impacts of an AI technology. The generated text is based on prompting GPT-3.5-turbo-16k with the illustrated prompt templates after replacing the placeholder text \texttt{\{Article\}} with the text of the article published by CNBC titled: \textit{These are the tech jobs most threatened by ChatGPT and A.I.} \cite{cnbc2023techjobs}.}
\end{figure*}\label{figure1}

Emerging Artificial Intelligence (AI) technologies can have a detrimental impact on individuals and society \cite{weidinger2023sociotechnical}, though negative impacts can sometimes be mitigated if taken into account early in the design process \cite{DBLP:journals/corr/abs-2108-07258, tamkin_understanding_2021}. One way to do so is through anticipatory approaches \cite{Brey:2012go} which seek to enumerate a range of plausible risks and consequences of a technology and how it may impact society, and then orient stakeholders to take responsibility for averting potential bad outcomes \cite{vesnic-alujevic_societal_2020, mittelstadt2015shape, diakopoulos2021nms}. Anticipating the negative impacts of a fast-moving emerging technology such as AI is difficult for a variety of reasons including the complex entanglements and interaction effects with the social worlds of human behavior and policy. Doing it effectively requires a deep understanding of an AI system's development life cycle including design, training, and deployment \cite{solaiman_evaluating_2023}. It also demands a robust mechanism to foresee a wide range of plausible impacts and minimize the uncertainty around the unintended consequences of AI systems on people and institutions \cite{metcalf_algorithmic_2021, shelby_sociotechnical_2023}. 

Recent efforts towards anticipating the potential risks and harms of AI systems have often relied on experts' perspectives \cite{schuett_towards_2023, DBLP:journals/corr/abs-2108-07258,weidinger_taxonomy_2022, nanayakkara_unpacking_2021}. Yet inadvertent expert biases may influence the foresight process and can lead to a distorted view of the future \cite{bonaccorsi_expert_2020}. Large Language Models (LLMs) have also recently been explored for their potential ability to expand the range of negative impacts considered by AI researchers and developers \cite{bucinca_aha_2023}, though they also suffer from concerns about the nature and extent of the biases that may be captured by their training data and so reflected in the generated text \cite{10.1145/3597307,talat_you_2022}. Therefore, the use of LLMs to help anticipate negative impacts of AI raises questions around the overall quality of the generated impacts, what categories of negative impacts different large language models are capable of generating, and how generated impacts are distributed with respect to those impact categories. Evaluating these dimensions in a systematic way is a crucial step towards understanding whether LLMs can become a reliable tool in larger AI impact assessment and anticipatory governance approaches.

In this work we address these questions by exploring and evaluating in greater depth the potential of LLMs (both fine-tune and zero-shot prompted) as tools to generate negative impacts of AI. First, we use GPT-3.5-turbo-16k to summarize the negative impacts that could be caused by AI systems from a large and diverse corpus of descriptions of AI impacts drawn from the news media. Although news media reflects its own set of normative biases about what is selected for coverage and how AI is covered \cite{10.1145/3306618.3314285,nguyen2022news}, by creating a diverse sample we use it as a baseline that captures a broad range of societal concerns about AI technologies. In addition, this corpus provides a ready source of material that can be used for evaluation purposes in terms of whether certain impacts can be generated based on the technical and contextual description of an AI system presented in a news article. We then evaluate the capabilities of several models for generating impacts with respect to a sample of articles, examining both zero-shot prompting and fine-tuning to assess how different approaches and models may shape the categories and distribution of impacts generated. More specifically, this research starts by addressing questions about the quality and categories of impacts generated by LLMs by qualitatively evaluating the negative impacts generated by closed-source (e.g. GPT-4) and open-source (e.g. Mistral-7B-Instruct) instruction-based models. We also extend this evaluation to assess whether completion models ( GPT-3 and Mistral-7B) fine-tuned on negative impacts from news coverage of AI across 266 online news domains in 30 countries can capture a more comprehensive range of categories of negative impacts when compared to impacts generated using instruction-based models.

Based on our analysis, we make three main contributions. First, qualitative assessment of generated impacts across the four dimensions of coherence, granularity, relevance, and plausibility shows that smaller open-source models such as Mistral-7B fine-tuned on negative impacts from news media tend to generate negative impacts that are qualitatively comparable to those generated using a larger scale model such as GPT-4. Second, our findings suggest that fine-tuning smaller open-source LLMs like Mistral-7B, with very limited compute resources, on a diverse data source such as the news media can cover a range of categories of negative impacts relevant to AI technologies beyond the ones anticipated using a much larger and capable LLM such as GPT-4. In addition, we highlight a potential bias in state-of-the-art LLMs when used for anticipating impacts. Specifically, both closed-source and open-source instruction-based models tend to miss categories of negative impacts that were generated by LLMs fine-tuned on negative impacts from the news media. And finally, these findings contribute to research on anticipatory ethics and governance \cite{DBLP:journals/corr/abs-2011-13170} by paving the way to broaden access to scalable tools that could be used in supporting the anticipation of negative impacts of AI in society. Moreover, the qualitative dimensions of quality and the benchmarks of performance across different models that we present may help to orient future development of such tools.

\section{Related Work}
Anticipating AI impacts helps practitioners and domain experts articulate, and potentially mitigate, the social and ethical implications of emerging technologies based on their contextual use and functional capabilities and affordances \cite{metcalf_algorithmic_2021, DBLP:journals/corr/abs-2011-13170}. While it is difficult to consider all potential impacts of new technologies due to uncertainty \cite{DBLP:journals/corr/abs-2011-13170}, various methodological approaches have been proposed to enable practitioners to anticipate and foresee failures and potential impacts of technologies that have not yet been observed or that may occur in new contexts.

Expert involvement in the anticipatory process has been considered to be essential in some anticipatory methods \cite{solaiman_evaluating_2023, bolger2017use}. For instance, researchers proposed co-designing a checklist with practitioners to help identify and mitigate against potential issues of the systems that are under development \cite{10.1145/3313831.3376445}. However, such checklists have been criticized for being broad and failing to consider differences between technologies, applications, and stakeholders as part of the evaluation process \cite{kiran2015beyond}. Expert peer review is another approach that has been explored by grant-awarding entities \cite{bernstein2021} and academic conferences such as NeurIPs \cite{nanayakkara_unpacking_2021} requesting researchers to include an impact assessment of their work or write about the potential impacts of the technologies they are researching. Some research argues that such initiatives may not be the most effective way to assess the impact of AI systems \cite{holbrook2011peer} because they may suffer from inadvertent expert bias, demographically skewed backgrounds, and homogenous experiences of experts \cite{bonaccorsi2020expert, crawford2016artificial}. 

More recent efforts in participatory ethics have leveraged the cognitive diversity of the crowds recruited through Amazon Mechanical Turk in anticipating the social impacts of automated and algorithmic decision-making systems \cite{barnett_crowdsourcing_2022}. However, research has found that the perception and understanding of AI, as well as the social impacts around its use, tend to be markedly different across geographical settings \cite{hagerty_global_2019, jobin_global_2019}. Accordingly, while participatory approaches involving crowdsourcing impacts provide a more diverse alternative to expert opinions in anticipating impacts, they still overlook the cultural and socio-technical factors of the crowds which are likely to reflect only the types of impacts deemed relevant and important to the culture and country the crowds belong to. Despite the efficacy of the growing methods in anticipatory and participatory foresight in AI ethics, many of these methods, such as the ones described earlier, do not directly assist AI systems designers and developers with the most accessible user experience to brainstorm and explore the range of impacts of AI technologies earlier in the development life cycle and before the deployment of these technologies. 

In response to some of these challenges, researchers have begun to explore the use of Large Language Models in the anticipatory process \cite{pang2024blip}. For instance, AHA! (Anticipating Harms of AI) is a generative framework that leverages GPT-3 to automate the generation of vignettes to elicit examples of the potential harms of AI on stakeholders based on the description and set of problematic behaviors provided by users \cite{bucinca_aha_2023}. In general, if such approaches are to be viable and gain use in anticipatory governance approaches, more evaluation is needed to assess the overall quality of the results. In this work, based on prior evaluation studies of generated text and of quality criteria applied in other anticipatory approaches \cite{Uruena:2019di, diakopoulos2021nms, sun2020improving, howcroft2020twenty} we develop a set of qualitative criteria related to impact coherence, granularity, relevance, and plausibility that help to evaluate and articulate efficacy. Moreover, LLMs have biases based on potential unbalanced and selection biases that exist in the datasets they are trained on \cite{10.1145/3597307, talat_you_2022} which could potentially influence the scope of impacts they generate.
Accordingly, in the absence of domain-specific benchmarks, it is crucial to assess the efficacy of LLMs in anticipatory impact assessment exercises based on the quality and range of categories of impacts these models are capable of generating to support a diverse set of outputs.

\section{Data Collection and Processing}

We first develop a corpus of news media reporting on AI systems in society to support our evaluation study. This corpus serves both as a diverse sample of AI systems and their impacts as reported around the world, as well as a dataset that we use to fine-tune models to evaluate their efficacy in larger AI impact assessment tasks. Although news media embeds its own set of biases related to what systems and impacts are selected for amplification to the broader public \cite{diakopoulos2024prospective}, by creating a diverse sample of media from a plurality of sources and parts of the world we hope to at least somewhat mitigate this issue. Ultimately though, news media is one of several possibilities for defining a baseline to study the capabilities of LLMs for impact generation, and future work may also consider other reasonable corpora. We describe in the following sections the process for curating this diverse corpus of news media.

\subsection{Curating AI-relevant keywords from news media}
To construct our dataset of a diverse sample of news articles covering AI and its negative impacts, we first developed a list of keywords relevant to AI technologies and applications to help focus our search. We started by using the New York Times (NYT) developer search and archive API endpoints\footnote{https://developer.nytimes.com/docs}, to retrieve 612 articles on AI published by the NYT between January 2017 and May 2023, inclusive, based on two broad seed search words (``A.I.", and ``Artificial Intelligence"). We then scraped the text for each item and extracted a list of n-grams (uni-gram, bi-gram, and tri-gram) which indicate commonly used words for covering AI technologies in the NYT. By manually selecting the n-grams relevant to AI and AI technologies, we identified a list of 31 relevant keywords spanning numerous topics relevant to AI. To further expand the comprehensiveness of the curated list of AI-relevant keywords, we also scraped the full text of 2,724 articles associated with 529 incidents between January 2017 and June 2023 that are relevant to AI from the AI Incident Database, which curates news items and other reports indicating AI failures in the real world\footnote{https://incidentdatabase.ai} \cite{mcgregor_preventing_2020}. Using the same n-gram extraction method mentioned earlier, we identified nine new keywords that didn't overlap with the 31 already found. This brought the total number of the curated keywords up to 40. A full list of the keywords is provided in the appendix (\ref{a.1}). Next, we describe how these keywords were used to scrape news articles relevant to AI. 

\subsection{Scraping \& Filtering News Articles}
We used Google News to search and retrieve all articles published in English based on the curated set of AI-relevant keywords. For each keyword in the curated list, we sent a search request to Google News via client URL (cURL) containing the search keyword and date range (between January 1st, 2020 and June 1st, 2023) to pull articles from. Using this method we retrieved the URL, title, and domain for each news article. The total number of retrieved articles from Google News based on the curated list of keywords is 665,965 articles. We successfully scraped full-text content for 89.4\% (595,371) out of 665,965 retrieved articles. Out of those, 240,850 (approximately 40.45\%) articles from 11,980 domains had at least one exact match to our list of keywords. 

We further focus this dataset on the top 5\% (402) of domains in our sample that published at least 100 articles or more. We observed that not all 402 domains are from news media. Therefore, we excluded from our dataset all non-media domains such as government agencies, academic institutions, or enterprise blogs. As a result, the final dataset consists of 91,930 articles that were published by 266 domains between January 1, 2020 and June 1st 2023. To determine the country of origin for the domains, as a proxy to the diversity of our sample, we visited the website for each of the 266 domains to check the physical address or city that is stated in the footer or the about page. News articles in our sample span 30 countries around the world, however the majority of news articles (84\%) were published by 10 countries: US (37,056), India (22,104), UK (8,543), Canada (2,480), China (1,815), Australia (1,541), UAE (1,186), Israel (1,095), Germany (770), and Turkey (668). In addition, 19.1\% (17,590) out of 91,930 articles covering AI in our sample discuss or mention negative impacts of AI, which is in line with previous work showing that the benefits of artificial intelligence are discussed more frequently in news media than its risks \cite{10.1145/3306618.3314285}.% 

\section{Methodology}
This section describes the methodology for generating and evaluating negative impacts. We first use an LLM to summarize two parallel pieces of information from each news article in our dataset: a description of the AI systems reported on, and a set of negative impacts described that are associated with these systems (as illustrated in Figure \ref{figure1}). We then describe how we generate impacts using this information through zero-shot prompting of instruction-based models (GPT-4 and Mistral-7B-Instruct). Afterwards, we elaborate on how we fine-tuned two completion models (GPT-3 and Mistral-7B) for the down-stream task of generating negative impacts using a curated sample from news media. Next we assess the capabilities of the models for generating negative impacts by evaluating the 1) quality and 2) coverage of the generated impacts to those present in our sample. To evaluate the quality of impacts, we describe a qualitative rubric we developed and employed to verify whether the generated impacts using LLMs are indeed negative and subsequently evaluate the quality of these impacts based on four dimensions: coherence, granularity, relevance, and plausibility. In order to evaluate the coverage of impacts generated by the models, we describe how we created a typology of impacts based on our sample from news media. We then detail how we used this typology as a baseline for comparing the range of impacts that the models generated with those present in this baseline.

\subsection{Summarizing negative impacts and descriptions of AI technologies}\label{4.1}

\renewcommand{\thetable}{S\arabic{table}}
\setlength{\textfloatsep}{0pt}
\begin{table*}[ht]
\centering
\footnotesize
    \begin{tabular}{|p{0.3cm}|p{8cm}|p{8cm}|} 
    \cline{2-3}
    \multicolumn{1}{c|}{} & Prompt & Description \\    
    \hline
    P1\label{p1} & Summarize the negative impacts explicitly mentioned in the following article. If no impacts are mentioned type only: \#\#No Impacts\#\#. \#\#\#Article\#\#\#: \{\textit{Article}\} & Prompt to extract the negative impacts of AI that are explicitly mentioned in the news articles using GPT-3.5-turbo-16k\\
    \hline
    P2\label{p2} & In a single paragraph, explain the functional capabilities of the technology described in the article, domain of use, stakeholders, and users without mentioning any negative aspects or concerns. Focus solely on the technology's features and its relevance to stakeholders and users. Be accurate. Do not make up information not described in the article. Let's think step by step. \#\#\#Article\#\#\#: \{\textit{Article}\} & Prompt to extract functional capabilities and the contextual use of AI technologies using GPT-3.5-turbo-16k \\
    \hline
    P3\label{p3} & You are given a functional description of a technology delimited by \#\#Description. \#\#Description: \{\textit{functional\_description\}}. Write a single negative impact of this technology based on the provided functional description. Limit your answer to one sentence. & Prompt for zero-shot generation of negative impacts using GPT-4. The prompt is formulated to include the functional and contextual descriptions of an AI technology and an instruction to generate a single negative impact of this technology based on the provided descriptions. \\
    \hline
    P4\label{p4} & \textless s\textgreater [INST] Describe a single negative impact of the technology described below and delimited by \#\#Description: \#\#Description \{\textit{functional\_description\}} Write a single negative impact of this technology based on the provided functional description. Limit your answer to one sentence. [/INST]\textless /s\textgreater & Prompt for zero-shot generation of negative impacts using Mistra-7B-Instruct. The prompt is formulated to include the functional and contextual descriptions of an AI technology and an instruction to generate a single negative impact of this technology based on the provided descriptions.\\
    \hline
    \end{tabular}
    \caption{Prompts templates used to a) extract the functional and contextual descriptions of AI technologies and their negative impacts from the news media and b) assess the proficiency of GPT-4 and Mistral-7B-Instruct instruction-based models in generating negative impacts using zero-shot prompting based on the descriptions of AI technologies in the test dataset. The text in curly brackets is replaced by the text collected or generated from the news media.}\label{tab:table8} 
\end{table*}

We chose GPT-3.5-turbo-16k to generate a summary of the negative impacts of AI that are mentioned in the articles. Prior work found that LLM summaries tend to be on par with human written summaries \cite{10.1162/tacl_a_00632}. Our choice of using GPT-3.5-turbo-16k was driven by the lack of classifiers trained to discern the types of impact descriptions of AI mentioned in news articles and the possibility of having multiple impacts described in each article.  In addition, the model’s large context window enabled us to embed the full article text in the prompt context. 

Prompt P1 in Table \ref{tab:table8} shows how we prompted the model to extract negative impacts from articles. We validated the prompt on a random set of 20 articles from our sample of articles and found that all outputs included one or more negative impacts related to the technology reported in the article. The model will indicate if an article does not state or discuss negative impacts in its text and those articles without negative impacts extracted were excluded from further analysis as our research is primarily oriented towards understanding negative impacts. In cases where multiple impact descriptions are extracted from a single article, each impact description receives an entry in our dataset linking it to the article it was extracted from.

We further used the model to generate descriptions of the functional capabilities and contextual use of AI technologies described in each news article. This allows us to build a dataset that connects these functional descriptions to the impact descriptions and facilitates later fine-tuning. This approach also creates an opportunity to translate documents from other sources into this intermediate representation before inferencing negative impacts, thus facilitating a wider array of potential downstream deployment contexts. 

Similar to the methodology followed in extracting negative impacts, we employ GPT-3.5-turbo-16k to prompt it for the description of the functional capabilities of the technology as mentioned in each article, as well as contextual aspects of the technology’s domain of use, stakeholders involved, and potential users of the technology who might be negatively impacted by the technology (see Appendix \ref{a.3} for examples). To improve the quality of the generated text and reduce hallucinations, we set the model temperature to 0.1 to be more deterministic and also instructed the model to do some reasoning before generating a response by thinking step-by-step following the proposed Chain of Thought (CoT) prompt structure \cite{kojima2023large}. The final prompt template shown in P2 in Table \ref{tab:table8} was used to prompt the model to describe functional capabilities and the contextual use of AI technologies. We also validated our prompt on the same sample of articles used to validate the prompts used to extract the negative impacts of AI and found the model to provide accurate summaries of the functional descriptions and mention of stakeholders involved as reported in the news articles without hallucinations (i.e., introducing additional details not mentioned in the article).

After extracting and synthesizing the descriptions of the technologies and their contextual use as mentioned in each article, we organized the dataset so that each description of a technology is paired with each negative impact for that technology that is identified in the same article. Curating the dataset in this format enables us to fine-tune models to generate a single impact description based on a functional description of an AI technology. The motivation behind this design decision is to offer future users a more extensible and flexible model that can be run multiple times to generate as many negative impacts as they desire per functional description of an AI technology.

The final dataset includes 37,689 pairs of descriptions and negative impacts of AI technologies from 17,590 articles. This curated dataset was split randomly into training (N=32,035), validation (N=5,140), and testing datasets (N=514). The training and validation datasets were used for fine-tuning and the testing dataset was used to evaluate the fine-tuned models in an impact generation task. We decided to keep the training sample large in order to not introduce additional biases in the selection of impacts used for fine-tuning and to preserve the diversity of impacts in the sample. Also, we chose a small testing sample to make the qualitative assessment of generated impacts more feasible.

\subsection{Zero-shot generation of negative impacts}\label{4.2}
To assess the proficiency of models for generating negative impacts, we prompted closed-source (GPT-4) and open-source (Mistral-7B-Instruct) instruction models using zero-shot prompting to generate negative impacts based on the descriptions of AI technologies in the test dataset. We selected the Mistral-7B \cite{jiang2023mistral} model for its small parameter scale and strong performance that matches the performance of much larger models such as LLaMa-13B \cite{touvron2023llama} on MT benchmark and LLaMa-34B on several NLP tasks. Moreover, it is more feasible to run inference on Mistral-7B-Instruct than a LLaMa-13B model with limited computing resources such as Google Colab.  

To generate a negative impact using instruction-based models, we used a zero-shot prompting technique \cite{brown2020language} – a type of prompting that does not necessitate the inclusion of examples of tasks in the prompt’s context. To use OpenAI's GPT-4 model for generating a negative impact, we formulated the prompt for this task as shown in P3, in Table \ref{tab:table8}, to include a description of the functional capability and contextual use of a technology and an instruction to generate a single negative impact of this technology based on the provided descriptions.\footnote{Besides limiting the maximum number of generated tokens by GPT-4 and Mistral-7B-Instruct through setting the \textit{max\_tokens} parameter in the OpenAI API and \textit{max\_length} in the text generation function in HuggingFace to 25, we also specified in the prompt to limit the generated negative impact to a single sentence. This was done to match the structure of the dataset we created. We also re-formatted the prompt to match the expected formatting of Mistral-7B-Instruct prior to generating the impacts. Each functional description was pre-appended by \textless s\textgreater [INST] and post appended by [/INST] (check prompt P4 in Table \ref{tab:table8}). These sequences inform the models where the instruction prompt starts and where it ends. To generate the negative impacts using Mistral-7B-Instruct we used HuggingFace text-generation pipelines and ran the model for inference on an A-100 GPU notebook instance on Google Colab Pro+.} 

\subsection{Fine-tuning LLMs to generate negative impacts}\label{4.3}
We fine-tuned two completion models (OpenAI GPT-3 and Mistral 7B) on the training dataset to further gauge the capability of LLMs in generating negative impact description without the need for prompt engineering. The selection of GPT-3 and Mistral-7B completion models for fine-tuning was also due to the lack of instruction datasets \cite{wang2023selfinstruct} focused on negative impacts of AI technologies. In addition, curating a representative dataset for instruction-tuning similar to the Alpaca dataset \cite{alpaca} but for negative impacts of AI falls out of the scope of this research as it requires crowdsourcing and validating representative seed questions from experts and practitioners that are deemed essential in the anticipatory process of AI impacts.

To fine-tune GPT-3 we first prepared the training and testing dataset according to the prompt-completion format following OpenAI’s legacy fine-tuning guide\footnote{https://platform.openai.com/docs/guides/fine-tuning/create-a-fine-tuned-model}. Each description of an AI technology was used as a prompt and the negative impact corresponding to that description was provided as the completion.\footnote{Functional descriptions and negative impact descriptions in the datasets were appended with \texttt{\textbackslash n\textbackslash n\#\#\#\textbackslash n\textbackslash n}
 and \texttt{"END"} to inform the model where the prompt and completion end, respectively. Using OpenAI API, GPT-3 was fine-tuned using OpenAI’s auto-allocated hyperparameters as recommended by OpenAI's fine-tuning guide. The model was trained for 2 epochs, on a batch size of 38, and learner multiplier is equal to 2.}

To fine-tune the Mistal-7B model, we reformatted the data to fit Mistral’s prompt-completion format. Descriptions of AI technologies and their corresponding negative impacts were prepended by \texttt{\#\#\#Input}, and \texttt{\#\#Response}, respectively. We attempted to match the fine-tuning hyperparameters of Mistral-7B to match GPT-3 to have a fair comparison of the output of both models. Accordingly, we fine-tuned Mistral-7B through QLoRA \cite{dettmers2023qlora}.\footnote{Fine-tuning of Mistral was done in Python on a Google Colab Pro+ using an A100 GPU and took approximately 4 hours. We fine-tuned the mode for 2 epochs, with a learning rate equal to 2e-5, and a training batch size of 32. The bitsandbytes library was used for quantization and PEFT for parameter efficient fine-tuning.} 

\subsection{Evaluating the generated impacts from LLMs}\label{4.4}
We developed a rubric to guide the qualitative assessment of the quality of the generated impacts across models (See Appendix \ref{a.2}). This rubric includes (1) verification in terms of whether the generated negative impacts are indeed negative impacts, and (2) the quality of the generated impact based on four dimensions: \textit{coherence}, \textit{granularity}, \textit{relevance}, and \textit{plausibility}. We selected these four dimensions as criteria for evaluating generated impacts by adopting the perspective of an end-user. We assume that users may prioritize the clarity, descriptiveness, alignment with the functional description used as input to the models, and reasonableness of the generated negative impacts using these models. In future work, we plan on extending the evaluation criteria to include domain-specific dimensions of importance that might be of interest to downstream users such as journalists, scientists, or policy-makers. 

Coherence is used to capture the comprehensibility of a generated impact in terms of complete sentences and the mention of a single impact in the anticipated impact using LLMs. Granularity describes the extent to which the generated impacts are sufficiently descriptive. Relevance is defined in terms of the relevance of the negative impact description to the functional capabilities of an AI technology and the entities involved in using it. The plausibility \cite{URUENA201915} dimension evaluates the reasonableness to conclude that the generative impact could happen given what is known about the socio-technical world. \

In order to evaluate the coverage and potential alignment of the generated impacts with respect to the types of impacts in our sample of news media, we first establish a data-driven typology that we can use to compare the categories and distributions of generated impacts. To do this we categorize the impacts in the full dataset by applying BERTopic \cite{grootendorst2022bertopic}, a topic modeling technique that leverages transformers to create easily interpretable topics.\footnote{To run BERTopic we configured the min\_topic\_size and zero\_shot\_min\_similarity to 10 and 0.9, respectively setting the the minimum number of sentences in a topic to be low and have a higher confidence score in assigning a topic to sentences.} The resulting set of ten topics were then manually labeled based on the keywords and three representative examples for each topic. Finally, we mapped the manually labeled topics back to the the negative impact descriptions represented in our sample. In the results section, we describe the categories of impacts emerged from our dataset and then use these categories to assess the coverage of impacts generated by the models.

\section{Results}\label{results}
\renewcommand{\thetable}{S\arabic{table}}
\begin{table*}[ht]
\scriptsize % Add this line to make the font even smaller
\centering
\begin{tabular}{@{}lp{4cm}ccccc@{}}
\toprule
Criterion & Description & Qualitative Rubric & GPT-4 & Mistral-7B-Instruct  & GPT-3 & Mistral-7B \\
\midrule

% validation of negative impacts
\multirow{2}{*}[1ex]{Validation} & Is the generated text a negative impact? & No & 0 (0\%) & 0 (0\%)  & 75 (14.59\%) &  47 (9.14\%) \\
& & Yes &  514 (100\%) & 514 (100\%) &  439 (85.40\%)  &  467 (90.85\%) \\
\midrule

% Coherence
\multirow{2}{*}[1ex]{Coherence} & Is the generated impact a complete sentence? & No & 0 (0.00\%) & 36 (7.00\%)  & 27 (6.15\%) & 21 (4.49\%)\\
& & Yes &  514 (100\%)& 478 (93.00\%)  & 412 (93.84\%) & 446 (95.50\%)\\
\midrule
\multirow{3}{*}[2.5ex]{Coherence} & Does the generated impact include more than one impact & No & 497 (96.69\%) & 462 (89.88\%)  & 395 (89.97\%) & 436 (93.36\%) \\
& & Yes &  17 (3.30\%) & 52 (10.11\%)  & 44 (10.02\%) & 31 (6.63\%)\\
\midrule

% Granularity
\multirow{3}{*}[2.5ex]{Granularity} & How elaborative is the generated impact? & Too concise & 0 (0\%) & 1 (0.19\%)  & 4 (0.911\%) & 7 (1.49\%) \\
& & Could explain more &  407 (79.18\%) & 320 (62.25\%) & 378 (86.10\%) & 381 (81.58\%)\\
& & Sufficient & 107 (20.81\%) & 193 (37.54\%)  & 57 (12.98\%) & 79 (16.91\%)\\
\midrule

% Relevance
\multirow{3}{*}[2.5ex]{Relevance} & How relevant is the impact to stakeholders? & Irrelevant & 2 (0.39\%) & 24 (4.66\%) & 4 (0.91\%) & 11 (2.35\%) \\
& & Somewhat Relevant &  29 (5.64\%) & 74 (14.39\%)  & 59 (13.43\%) & 20 (4.2\%)\\
& & Very Relevant &  483 (93.96\%) & 416 (80.93\%) & 376(85.65\%) & 436 (93.36\%)\\
\midrule
\multirow{3}{*}[2.5ex]{Relevance} & How relevant is the impact to the functional capabilities of the technology? & Irrelevant & 13 (2.53\%) & 22 (4.28\%) & 12 (2.73\%) & 19 (4.06\%) \\
& & Somewhat Relevant &  114 (22.17\%) & 83 (16.14\%) & 53 (12.07\%) & 37 (7.92\%)\\
& & Very Relevant &  387 (75.29\%)& 409 (79.57\%) & 374 (85.19\%) & 411 (88.00 \%)\\
\midrule

% Plausibility
\multirow{3}{*}[2.5ex]{Plausibility} & How plausible is the generated impact? & Not Plausible & 0 (0.00\%) & 0 (0.00\%) & 0 (0.00\%)  & 0 (0.00\%)\\
& & Somewhat Plausible &  3 (0.58\%) & 10 (1.94\%) & 38 (8.65\%) & 20 (4.28\%)\\
& & Very Plausible & 511 (99.41\%) & 504 (98.05\%) & 401 (91.34\%) & 447 (95.71\%)\\
\bottomrule
\end{tabular}
\caption{Results of the qualitative evaluation of the generated impact statements on Coherence, Granularity, Relevance, and Plausibility using instruction-based and fine-tuned Large Language Models. The percentages denote the proportion of negative impacts satisfying each rating of the evaluation dimensions to the total number of negative impacts generated by each respective model.}
\label{tab:table5}
\end{table*}

In this section, we begin by describing our findings from verifying whether the generated impacts are negative and conducting a qualitative evaluation of these impacts according to the established criteria in section \ref{4.4}, namely \textit{coherence}, \textit{granularity}, \textit{relevance}, and \textit{plausibility}. We then move to share a set of categories of negative impacts emerging from our news articles dataset. Furthermore, we elaborate on our findings from using these categories to assess the performance of the models with respect to the \textit{distribution of impacts} generated by prompting instruction-based and fine-tuned models and those reflected in our test sample. In the following sections, our findings further support the capability of LLMs, especially if fine-tuned, in anticipating the negative impacts of AI technologies.

\subsection{Evaluating the quality of generated impacts}
All of the impacts generated by the models we evaluated were first verified as negative impacts (see Appendix \ref{a.2}). Impacts generated by instruction-based LLMs were all negative and the majority of the generated impact statements by the fine-tuned models were also deemed negative (see Table \ref{tab:table5}) as per the evaluation rubric. Accordingly, the remaining evaluation criteria (coherence, granularity, relevance, and plausibility) were applied only to the set of \textit{negative} impact statements generated by each model. 

%Next, we qualitatively describe our findings by analyzing the generated negative impacts in terms of coherence, structure, relevance, and plausibility. 

In terms of \textit{coherence}, the majority of the generated impacts using instruction-based and fine-tuned models were complete sentences but a small percentage of the generated negative impacts by each LLM had more than a single negative impact in the generated text (see Table \ref{tab:table5}). 

As for \textit{granularity}, the anticipated negative impacts were evaluated to be slightly abstract and could be elaborated on further. For instance, for this generated impact using GPT-3 ``the use of AI to manipulate and persuade voters can lead to a loss of trust in politics and democracy", a more complete generated impact would elaborate further on \textit{how} the use of AI would manipulate the voters and influence the voters' trust in politics and democracy similar to the impact generated by GPT-4: ``AI could undermine the democratic process by manipulating voters' choices based on targeted and potentially misleading information". 

For the \textit{relevance} criterion, the generated text was predominantly evaluated to be relevant to the functional and contextual descriptions of AI and to impacted stakeholders.  

Lastly, both instruction-based and fine-tuned models were able to generate somewhat or very \textit{plausible} negative impacts. For example, using a functional and contextual description sourced from the news media on NeuraLink Technology aiming to implant a chip in the skull to allow individuals with disabilities to regain the ability to walk, see, and communicate, Mistral-7B generated a somewhat plausible negative impact on ``the potential for misuse of Neuralink technology for mind control". In comparison, GPT-4 generated a very plausible negative impact pertaining to ``the risk of privacy violations, as personal thoughts and experiences could potentially be accessed and misused if the technology is hacked".

The qualitative evaluation of the generated negative impacts on coherence, granularity, relevance, and plausibility demonstrates the potential of smaller and open-source LLMs, such as Mistral-7B, to generate negative impacts that are relatively similar in quality to negative impacts generated by larger scale LLMs such as GPT-4 (see  Table \ref{tab:table5}). 

\subsection{Evaluating the coverage of generated impacts}
\begin{table*}[ht]
    \centering
    \centering
\footnotesize % Add this line to make the font even smaller
\begin{tabular}{ l c  c  c  c  c c  c  c}
\toprule

Impact Category & Test dataset & GPT-4 & Mistral-7B-Instruct & GPT-3 & Mistral-7B \\
\midrule
Societal Impacts & 42.02\% & 26.65\% & 25.29\% & 35.99\% & 41.75\% \\
Privacy & 16.53\% & 23.73\% & 16.92\% & 12.98\% & 9.85\% \\
Economic Impacts & 9.92\% & 24.51\% & 33.46\% & 8.88\% & 13.91\% \\
Accuracy and Reliability & 7.19\% & 9.33\% & 7.00\% & 11.16\% & 8.77\% \\
\hline
AI Governance & 7.19\% & 0.77\% &\colorbox{lightgray}{0.00\%} & 9.11\% & 6.42\% \\
\hline
Miscellaneous Impacts & 6.42\% & 3.69\% & 8.36\% & 9.56\% & 7.70\% \\
Physical and Digital Harms & 5.25\% & 7.78\% & 4.66\% & 4.10\% & 7.06\% \\
Security & 2.33\% & 3.50\% & 3.50\% & 4.78\% & 0.64\% \\
\hline
AI-generated Content & 1.94\% & \colorbox{lightgray}{0.00\%} & \colorbox{lightgray}{0.00\%} & 1.13\% & 0.85\% \\
Autonomous System Safety & 1.16\% & \colorbox{lightgray}{0.00\%} & 0.77\% & 2.27\% & 2.99\% \\
\bottomrule
\end{tabular}
\caption {
Prevalence of the categories of negative impacts in the test dataset and the generated impacts using instruction-based (GPT-4 and Mistral-7B-Instruct) and fine-tuned completion (GPT-3 and Mistral-7B) models based on the  typology of impacts developed from our dataset (see Section \ref{5.2}). Generated impacts are based on the functional descriptions and contextual use of AI technologies in the test dataset. The percentages denote the proportion of negative impacts in each category to the total number of negative impacts generated by each respective model. The cells highlighted in gray indicate the categories of impacts missed by GPT-4 and Mistral-7B-Instruct.
} 
\label{tab:table6} 

\end{table*}
\subsubsection{Establishing a typology of impact categories}\label{5.2}
A total of 10 categories emerged from the negative impact statements in our sample relating to: Societal Impacts, Economic Impacts, Privacy, Autonomous System Safety, Physical and Digital Harms, AI Governance, Accuracy and Reliability, AI-generated Content, Security, and Miscellaneous Risks and Impacts. Next, we describe each category in more detail including some examples of each.

\textit{Societal Impacts} – The impacts in this category describe the social implications of misusing AI for malicious purposes such as “spreading misleading ideas”, “spread[ing] disinformation and erode[ing] public trust”, and “overhlem[ing] the democratic process through the massive spread of plausible misinformation through AI systems”. Moreover, AI-powered applications that create Deepfakes were also a prominent topic in this category surfacing social and ethical considerations beyond using the technology for ``coordinate[ed] misinformation campaigns" to include ``defamation and blackmailing" and the misuse of the technology to “defraud companies”. Additional impacts in this category also captured some biases reflected or exacerbated by AI such as misidentifying ``people of color and transgender and nonbinary individuals".

\textit{Economic Impacts} – This category describes the potential and realized impacts of using or deploying AI across industries. Impacts in this category discussed the potential of AI to cause ``economic uncertainty and job displacement" such as ``potential displacement of jobs due to AI powered chatbots". In addition, some impacts describe how the belief that AI ``can do most jobs" has ``caused job terminations in the tech industry".

\textit{Privacy} –  This category focuses on the potential privacy violations resulting from using, adopting, or deploying AI systems for monitoring and surveillance. In particular, this category is predominantly centered around describing the impacts of technologies such as facial recognition in “surveillance” and its ``potential use for harassment" which could undermine ``privacy and free speech” and ``poses a threat to civil rights".

\textit{Autonomous System Safety} – This category focuses predominantly on the negative implications of emerging technologies such as autonomous vehicles or drones on safety. An example of these impacts include the potential of autonomous vehicles to “cause crashes” or for drones to ``increase in civilian causalities" during warfare.

\textit{Physical and Digital Harms} – This category encompasses potential digital and physical harms caused by AI. Digital harms reflect the types of harms resulting from the cloud or online deployment of AI systems or technologies such as chatbots ``engaged[ing] in sexually explicit conversations with paying subscribers" or, in the context of facial recognition systems, the ``wrong conviction of black men due to incorrect facial recognition matches”. In contrast, existential threats and the impacts of AI in warfare focus on physical harms. Some articles focused on the potential threats of AI and ``artifical general intelligence (AGI)" on human life such as ``the destruction of humanity and the rule of robots" or the ``risk of someone losing their life due to an AI system's advice or action”. 

\textit{AI Governance} – This category describes the importance and need for setting up a regulatory framework to govern the development and deployment of responsible AI. This category also includes challenges in AI governance that are often framed as due to the “black box problem, where it is difficult to know when an AI is confident or uncertain about a decision” and to the lack of ``accountability for how they [AI systems] are built or tested". For instance, the ``lack of repeatability and interpretability in AI models” makes it difficult to ``explain and justify decisions made by generative AI system". Additional challenges include “update[ing] and align[ing] AI systems with democratic values such as fairness, privacy, and protection from [potential misuse for] online harassment and abuse".

\textit{Accuracy and Reliability} – 
This category describes concerns pertaining to the reliability of AI such as “overtrust[ing] robots and technology, leading to automation bias” or its “tendency to hallucinate information and generate false or misleading statements” that are ``plausible but incorrect". Moreover, LLM models such as ChatGPT raise concerns about their potential to ``create realistic content that appear accurate" without ``reveal[ing] the sources of its information" which deem them as ``unreliable for real life settings".

\textit{AI-generated Content} – The category portrays the challenges in detecting the different modalities (images, audio, and text) of AI generated content and the potential impacts of such content. For instance, the “difficulty in distinguishing fake images” is making the task ``more challenging for law enforcement to identify and rescue victims [of child pornography]". Additional impacts include problems in “distinguish[ing] real from an AI-generated voices” which has the potential to be misused beyond ``voice cloning scams" such as ``strip[ing] away a celebrity's agency" over their voices. Additional impacts of AI-generated content also includes the impacts of “AI generated text [that] may not be detectable by existing plagiarism software” on ``academic integrity”.

\textit{Security} –  This describes the methods and consequences of exploiting security vulnerability of AI technologies for malicious purposes. For instance, cybercriminals could exploit generative AI for ``cyberattacks", ``malware and ransomware", and ``phishing and fraud" leading to ``new and improved [cyber]attacks” using techniques such as ``prompt injection attacks”.

\textit{Miscellaneous Impacts} – This catch-all group includes all remaining negative impacts that raise other important negative consequences of AI, but were not prominent enough to be represented as their own categories. This included impacts such as the cost of training AI models like LLMs ``the cost of training AI models on large datasets is expensive” or environmental impacts because ``data centers supporting AI models contribute to carbon emissions". Also, negative impacts of AI on cognition such as ``information overload" due to ``GPT's ability to generate lot of text which makes it difficult to distinguish between fact and fiction" or impacts of AI chatbots on emotions such as ``inspire[ing] false feelings of requited love in vulnerable individuals".

\subsubsection{Evaluating the distribution of generated impacts}
All negative impacts in our datasets were generated once using LLMs for each functional and contextual description of a technology. Accordingly, this section describes results from our analysis of whether the negative impacts generated by fine-tuned and instruction-based LLMs cover the range of impacts present in the test dataset. By reviewing and manually labeling the generated negative impacts based on the 10 categories of impacts described  earlier, we provide an overview of the alignment between the categories of negative impacts generated using the LLMs and those found in our sample of articles from the news media.

We found that the generated negative impacts using instruction-based and fine-tuned models largely reflect the categories of negative impacts present in the test dataset, though with some gaps in the production of certain categories by the instruction-based models. Overall, the generated negative impacts by both fine-tuned models (GPT-3 and Mistral-7B) covered all 10 categories of negative impacts, whereas the GPT-4 model did not produce any impacts in the AI-generated Content or Autonomous System Safety categories and the Mistral-7B-Instruct model did not produce any impacts in the AI Governance or AI-generated Content categories (see Appendix \ref{a.4} for some examples). As shown in Table \ref{tab:table6}, the two categories of negative impacts that GPT-4 missed (AI-generated Content and Autonomous System Safety) account for approximately 3.84\% and 3.40\% of the total generated negative impacts by the fine-tuned models (Mistral-7B and GPT-3), respectively. Moreover, the category of impacts pertaining to AI Governance that is missed by Mistral-7B-Instruct accounted for 6.42\% and 9.11\% of the total generated negative impacts by Mistral-7B and GPT-3, respectively. These findings demonstrate the capability of LLMs fine-tuned on news media to anticipate a range of negative impacts that go beyond the zero-shot impacts generated using larger-scale and instruction-based LLMs.

Drilling further into these disparities, we find a higher degree of concentration in the impact types produced by instruction-based in comparison to fine-tuned models. To support our observation, we calculate the Gini coefficient based on the number of negative impacts generated using each LLM and present in the test dataset. By calculating the Gini coefficient, we are able to statistically compare how concentrated the unequal distributions of negative impacts are, with respect to categories of impacts, across models and compare these distributions to the distribution of impacts present in the test dataset. A model that has a high Gini coefficient (more unequal) has a higher concentration of impacts in some categories. In contrast, models with lower Gini coefficient (more equal) have a more even (i.e., spread out) distribution of negative impacts across the categories of impacts. Despite the minor differences in the coefficients calculated for each model, we find that Mistral-7B (0.50) and GPT-3 (0.44) tend to have a more even distribution of negative impacts across categories, and are closely aligned with the distribution of impacts in the test dataset (0.51), compared to GPT-4 (0.55) and Mistral-7B-Instruct (0.58) which are slightly more uneven and concentrated in terms of the categories of impacts generated.

%In the next section, we delve deeper into the analysis to evaluate how the categories of anticipated negative impacts vary between instruction-based and fine-tuned models.

% \begin{figure*}
% \centering
% \includegraphics[width=\textwidth]{AIES-impacts-submission/figures/categories_of_impacts.png} %
% \centering
% \caption{Place holder for impact categories}
% \end{figure*}

\section{Discussion}
This research presents an evaluation of the capabilities of LLMs in supporting an anticipatory impact assessment task concerning the generation of negative impacts based on the technical and contextual descriptions of AI technologies. To do so, we first curated a diverse corpus of descriptions of AI impacts and the functional descriptions and contextual use of AI technologies drawn from the news media. This dataset serves as a baseline that captures various types of impacts and a range of societal concerns about AI technologies. The emerging set of categories of impacts from this baseline were used as a reference to examine how different approaches (zero-shot prompting and fine-tuning) and models (instruction-based and completion models) may shape the categories and distribution of impacts generated with what we found from news media.

Our findings from evaluating the capability of instruction-based and fine-tuned LLMs for generating the impacts of emerging technologies demonstrate overall high levels of performance across metrics of coherence, granularity, relevance, and plausibility with relatively minor variations in performance across different models. The open-source Mistral-7B-Instruct model showed serviceable performance in comparison to the much larger and proprietary GPT-4, and when fine-tuning the completion model of Mistral-7B, it exhibited a more aligned distribution of impacts with respect to the news media, demonstrating the potential value of fine-tuning for this task. 

These findings also suggest that these models are already performant enough to be utilized by various stakeholders seeking to leverage LLMs to support anticipatory thinking about the impacts of AI. For example, fine-tuned models on descriptions of impacts could be integrated into the scientific process to help researchers think creatively about the broader impacts of their research \cite{nanayakkara_unpacking_2021} (see also our ethics statement for this paper which was enhanced using the techniques presented in this paper). Furthermore, technology companies could potentially deploy these models as a tool for inference that augments developers' abilities to ideate and think broadly about the negative and unintended impacts of the technologies they are creating early on in the development life cycle, perhaps as part of internal auditing and evaluation processes \cite{raji2020}. Also, models fine-tuned on descriptions of impacts have the potential to be integrated with user-friendly interfaces to enable domain experts with minimal technical experience, such as journalists and policymakers, to also think about the negative impacts of emerging technologies, potentially supporting creative identification of angles for further reporting \cite{petridis2023}, or informing risk and impact assessment methodologies called for by policies like the EU AI Act \cite{madiega2021artificial, commisionregulation}. 

Our findings further highlight a potential bias in state-of-the-art LLMs when used for generating impacts and demonstrate the advantages of aligning smaller LLMs with a diverse range of impacts, as reflected in the news media, to capture such impacts during anticipatory exercises. 

The categories of impacts in our sample capture many of the impact categories outlined by recent research on the harms, as well as social and ethical impacts of AI \cite{shelby_sociotechnical_2023, solaiman_evaluating_2023}. However, it appears that impacts relevant to alienation and loss of agency that are reported in the literature \cite{shelby_sociotechnical_2023,weidinger2023sociotechnical} are missing from the news media, at least at the level of detail considered in this work. Future work may be warranted to drill into sub-categories of impact in the news media to further assess patterns in media coverage of AI, such as biases in framing that could lead to missing out on such longer-term thematic treatments of impacts, or to better understand the prevalence of presumptively dominant narratives. For instance, reading into the \textit{Physical and Digital Harms} category of impacts in our sample we noticed very few instances of impacts discussing existential threats of AI, providing little evidence that news media has succumbed to the rhetoric of a doomsday scenario of AI. Therefore, future work focusing on creating typologies of impacts on the sub-category level may provide more granular insights and potential for analysis of the nature of impacts constituting each of the categories of impacts found in our taxonomy. 

It is important to be clear that any categorization of impacts relying on news media will reflect biases based on the sources of data used to build it and future work is still needed to parse the potential sources of bias in a news-based taxonomy, such as news outlet credibility (i.e., low vs. high credible news sources), political bias, temporal biases, geographic biases, and even the type of news article (i.e. hard news vs. opinion). While our research falls short of accounting for these biases, we suggest that future research should account for them and evaluate the level of influence these biases have on the categories of impacts prevalent in the news media and subsequently on the range and quality of impacts that might be generated using LLMs fine-tuned on that data. One benefit of categorizing impacts that leverage news media is that they have the potential to continuously adapt and evolve as rapid developments of AI are reported on by the media. Paired with an approach based on fine-tuning, this could help LLMs used for generating impacts stay more closely aligned with how the society (as reflected in news media) evaluates the technology. Still, the biases present in news media (or any other data source chosen to act as a basis for fine-tuning for this task) raises an important question of how such biases would come to be reflected in an anticipatory process. For instance, if a norm is established that more anticipatory attention should be given to environmental impacts from AI, perhaps a training set could be modified to project that impact more frequently (while maintaining relevance) in a fine-tuned model. The findings in this work make clear that close attention to the biases in an underlying fine-tuning dataset will be crucial to attend to, measure, and potentially deliberate on in order to make models viable contributors to anticipatory governance approaches.

\section{Conclusion}
Overall, this research offers a step towards democratizing the process of anticipating the impacts of emerging technologies by using LLMs and leveraging more accessible and diverse assessments of emerging technologies such as those present in the news media. More specifically, we evaluated the potential of LLMs for brainstorming and exploring the negative impacts of AI using news media as a baseline. To establish this baseline, we summarized a diverse corpus of descriptions of AI technologies and their impacts drawn from a sample from the news media to capture a broad range of societal concerns about AI technologies. Our findings from the qualitative assessment of generated impacts across the four dimensions of coherence, granularity, relevance, and plausibility shows that smaller open-source models such as Mistral-7B fine-tuned on negative impacts from news media tend to generate negative impacts that are qualitatively comparable to those generated using a larger scale model such as GPT-4. In addition, our findings from evaluating the categories of negative impacts generated using instruction-based (GPT-4 and Mistral-7B-Instruct) and fine-tuned models (GPT-3 and Mistral-7B), with respect to the baseline from news media, suggest that fine-tuning smaller open-source LLMs like Mistral-7B, with very limited compute resources, on a diverse data source such as the news media can cover a range of categories of negative impacts relevant to AI technologies beyond the ones anticipated using a much larger and capable LLM such as GPT-4. Our research highlights a potential bias in state-of-the-art LLMs when used for anticipating impacts and demonstrates the potential of aligning smaller LLMs with a diverse range of impacts, such as those reflected in the news media, to better reflect such impacts during anticipatory exercises.

\subsubsection{Acknowledgements\\} This work is supported by the National Science Foundation via award IIS-1845460.

\subsubsection{\emph{Ethical Statement\\}}\label{appendix.a}
% \newpage
% \section{Research Ethics and Social Impact}
This work paves the way for future research to build anticipatory tools to guide practitioners and researchers in the process of anticipating the negative impacts of AI technologies. Depending on the context of the deployment, a range of unintended consequences could influence users' trust and reliance on these anticipatory tools. For instance, over-relying on our fine-tuned models, if deployed as an inference tool, has the risk of diminishing critical thinking and the anticipation of negative impacts if the outputs of the models are perceived or deemed to be conclusive or inclusive of all plausible and possible scenarios in which an AI technology could be used. Accordingly, we view the development of anticipatory tools as supporting methods (but not substitutions) in the creative process of anticipating the negative impacts of AI. 

Concerning our use of news media as a data source, we recognize and acknowledge the sensitivity surrounding the online scraping of articles to train Large Language Models, including questions currently before courts related to the fair-use doctrine. As a result, we do not currently plan to share or re-distribute the news article dataset collected or the models fine-tuned for the purpose of this research for any commercial purposes.\

\subsubsection{Generating impacts of this research using LLMs}To extend the scope of impacts beyond the ones we have considered or thought of, we leveraged GPT-4 and Mistral-7B fine-tuned on impacts from the news media to assist us with capturing the range of potential unintended consequences of using LLMs for anticipating impacts as described in the context of this research. By prompting GPT-4 with the abstract to this paper and prompt P2\footnote{See Table \ref{tab:table8}}, we extracted the functional and contextual descriptions\footnote{``The article discusses the application of Large Language Models (LLMs) like GPT-3 and Mistral-7B to aid in anticipatory governance by generating insights into the potential impacts of emerging AI technologies on society. These LLMs are utilized to assist experts in identifying and exploring a range of possible adverse outcomes of new technologies, thereby facilitating informed decision-making and policy development. The research compares the effectiveness of various models, including instruction-based and fine-tuned LLMs, in producing coherent, relevant, and plausible outputs. It finds that smaller models, like Mistral-7B, which are fine-tuned on diverse news media articles, can generate impacts of similar quality to those produced by larger, more advanced models such as GPT-4. This capability provides a scalable method to forecast diverse social impacts, thus enabling proactive governance measures".} of LLMs for anticipating negative impacts. Then we included these descriptions as part of the context of prompt P3 and generated five negative impacts per model. Although GPT-4 anticipated impacts that were covered in our research, such as bias in training data\footnote{``One negative impact of using large language models for anticipatory governance could be the potential bias in generated insights, particularly if the models are predominantly trained on data reflecting specific cultural or societal norms, which may not accurately represent global perspectives".} and over-reliance\footnote{``The reliance on these models might reduce the involvement of human experts in policy-making, potentially leading to oversights or misinterpretations of complex social issues that AI does not fully comprehend".} on LLMs for impact assessment, it also extended the scope of impacts to include a crucial point about the potential impact of using such anticipatory technologies on the public perception and trust in governance policies\footnote{``Erosion of public trust in governance, as communities may perceive decisions influenced by AI as less transparent or accountable".}. Also, it generated an impact related to the potential misalignment of generated impacts by the models with the goals and requirements of policies leading to ineffective exploration of governance strategies\footnote{``Misalignment between the model outputs and actual policy needs, potentially leading to ineffective or inappropriate governance strategies".}. Similarly, Mistral-7B contributed novel impacts beyond what we had considered, having to do with the model hallucinating impacts that are not grounded in reality\footnote{``These LLMs can produce coherent, relevant, and plausible outputs, but they may also generate impacts that are not grounded in reality".} and the potential of anticipatory tools to generate false or misleading information about the potential impacts of emerging AI technologies\footnote{``The use of large Language Models for anticipatory governance may lead to the generation of false or misleading information about the potential impacts of emerging AI technologies".}. This raises a political point about the potential exploitation of such tools by adversaries to promote false perception of AI technologies or mislead the public opinion regarding the impacts of these technologies on society which may influence the public support for the ongoing collaborative efforts towards establish governing policies for AI.
\bibliography{main}

\newpage
\appendix
\newpage

\section {Appendix}
\subsection{AI-relevant Keywords}\label{a.1}
The set of keywords used to probe the news media for articles on AI:\\
A.I., Artificial Intelligence, Automated Decision Making, Automated System, Autonomous Driving System, Autonomous Vehicles, Autonomous Weapon, Chat Bot, Chatbot, ChatGPT, Computer Vision, Deep Learning, Deepfake, Driverless Car, Facial Recognition, General Artificial Intelligence, Generative AI, GPT, Image Generator, Intelligence Software, Intelligent Machine, Intelligent System, Language Model, Large Language Model, LLMs, Machine Intelligence, Machine Learning, Machine Translation, Natural Language API, Natural Language Processing, Neural Net, Neural Network, Predictive Policing, Reinforcement Learning, Self-Driving Car, Speech Recognition, Stable Diffusion, Synthetic Media, Virtual Reality, Weapons System.

\subsection{Examples of functional and contextual descriptions generated by LLMs}\label{a.3}

\begin{enumerate}[label=\thesubsection.\arabic*, ref=\thesubsection.\arabic*]
    \item \label{item:c1} C1: ``Text generating AI tools like OpenAI's ChatGPT have the capability to generate books in a matter of hours, making it easier for aspiring authors to quickly produce content. The domain of use for this technology is the book industry, specifically self-publishing platforms like Amazon's Kindle direct publishing. The stakeholders involved include authors, readers, and the literary ecosystem. Users of this technology are the authors who utilize AI tools to generate books and publish them on platforms like Kindle. The relevance of this technology to stakeholders and users is that it provides a faster and more accessible way to create and publish books, allowing authors to reach a wider audience and readers to have a broader selection of content to choose from."
    \item \label{item:c2} C2: ``Artificial Intelligence (AI). AI is a mechanism that abstracts and reacts to content like humans, designed and developed by humans. It has the capability to learn from interactions with users stakeholders and users of AI include individuals and organizations who rely on AI for various purposes, such as risk assessment in the legal system or chatbot interactions on social media platforms."
    \item \label{item:c3} C3: ``Artificial Intelligence (AI) and algorithms has the capability to automate various processes and decision-making tasks in different domains such as social media, criminal justice, healthcare, education, and hiring. the stakeholders involved include individuals, organizations, and institutions that rely on AI systems for various purposes. the users of this technology are individuals who interact with AI systems, such as social media users, job applicants, and individuals affected by algorithmic decision-making in areas like criminal justice and healthcare. the relevance of this technology lies in its potential to improve efficiency and decision-making."
    \item \label{item:c4} C4: ``Voice Deepfakes, which are synthetic voices that closely mimic a real person's voice, replicating tonality, accents, cadence, and other unique characteristics. this technology is relevant to stakeholders such as speech synthesis and voice cloning service providers like ElevenLabs, as well as users who utilize AI and robust computing power to generate voice clones or synthetic voices. the process of creating voice Deepfakes requires high-end computers with powerful graphics cards and specialized tools and software. research labs are using watermarks and Blockchain technologies to detect Deepfake technology, and programs like DeepTrace are helping to provide protection."
    \item \label{item:c5} C5: ``An example of a functional and contextual description of AI used in the prompt that generated a negative impact that was evaluated as irrelevant according to the Relevance dimension: "Artificial Intelligence (AI). Its functional capabilities include the ability to process large amounts of data quickly, identify potential forced or child labor in supply chains, improve crop rotation and yields, help catch poachers, and protect endangered species. AI has the potential to revolutionize and improve various fields, such as education, climate change, agriculture, and health. The stakeholders involved in AI include the United Nations (UN), member states, governments, public sector institutions, companies, and experts. the users of AI can be Governments in Africa and organizations working to protect endangered species."
    \item \label{item:c6} C6: ``Driverless cars are capable of operating without a human driver and are currently being tested in cities like San Francisco, Phoenix, Austin, and Los Angeles. stakeholders involved in this technology include General Motors cruise and Google sibling Waymo. the technology's functional capabilities include obeying traffic rules and driving at the speed. users of this technology are the general public who share the roads with driverless cars.".
    \item \label{item:c7} C7: ``Artificial Intelligence (AI). it has the functional capabilities to generate plausible responses to prompts from users in various formats, such as poems, academic essays, and software coding. it can also produce realistic images, like the pope wearing a puffer jacket. the relevance of AI to stakeholders, such as Google's parent company alphabet, is evident as they own an AI company called deepmind and have launched an AI-powered chatbot called bard. users of AI technology, including radiologists, writers, accountants, architects, and software engineers, can benefit from its capabilities in assisting with tasks and prioritizing cases"
    \item \label{item:c8} C8: ``Generative AI tools, specifically OpenAI's latest product called ChatGPT. This large language model (LLM) has the capability to generate coherent paragraphs of text and can be instructed to write about various topics, including science. the stakeholders involved in this technology are academic journal publishers, such as Science and Springer Nature, who have introduced new rules addressing the use of generative AI tools in their editorial policies. the users of this technology are researchers and academics who utilize ChatGPT to assist in writing their research papers. the relevance of this technology to stakeholders and users lies in its ability to generate text and aid in the writing process, potentially improving efficiency and productivity in academic research."
\end{enumerate}

\subsection{Examples of instances missed by LLMs}\label{a.4}
\begin{enumerate}
    \item {The AI-generated Content category portrays the challenges in detecting the different modalities of AI generated content and the potential impacts of such content. For example, by prompting our fine-tuned GPT-3 model with the functional and contextual description of ChatGPT from the news media ~\ref{item:c8} the model generated an impact related to the AI-generated content and the ``integrity" of academic research. Similarly, Mistral-7B, generated a negative impact pertaining to ``concerns about the authenticity of the AI generated content" when used in academic research. In contrast, using the same functional and contextual description, GPT-4 generated a negative impact relevant to the cognitive impacts resulting from the ``reliance on ChatGPT for academic writing [which] could lead to a decrease in critical thinking" without mentioning any potential impacts of AI-generated content. Likewise, the impact generated by Mistral-7B-Instruct focused on the over-reliance on ChatGPT in academic research which may lead to ``a decrease in the quality of research papers, as some researchers may rely too heavily on the tool" when conducting research. Additional negative impacts in this category that were generated by Mistral-7B and GPT-3 include: how AI generated content is becoming ``indistinguishable from human writing, making it difficult to detect" and  how ``AI generated text can mimic the style and structure of academic writing". Additional impacts include the use of ``ai-generated content..to spread misinformation and propaganda" and the challenges of AI-generated art work ``rais[ing] questions about the boundaries between ai-generated art and original artwork".}

    \item {With respect to the Autonomous System Safety category, when prompting the models to generate a negative impact of driverless cars ~\ref{item:c6} Mistral-7B-Instruct generated a negative impact similar in context to the impact generated by GPT-4 in terms of the potential ``loss of jobs for professional drivers, such as taxi and truck drivers, as the demand for human-operated vehicles decreases", whereas fine-tuned Mistral-7B generated an impact pertaining to the safety of driverless cars: ``driverless cars may not be as cautious as human drivers, leading to more accidents".}

    \item {For the AI Governance category, Mistral-7B generated an impact about the ``need for a global regulatory framework for AI to ensure safety and addresses concerns regarding the potential for AI to be used for malicious purposes such as creating fake news and spreading misinformation" when prompted about AI ~\ref{item:c7}. In contrast, using the same functional and contextual descriptions of AI, GPT-4 generated an impact about the potential misuse of ``AI's ability to generate plausible responses and produce realistic images [that] could potentially lead to the creation and spread of misinformation or fake news". Mistral-7B-Instruct also had a similar generated impact focusing on AI's capability to generate realistic videos that ``appear accurate but are actually fabricated". Other generated impacts by Mistral-7B include the impacts of ``the lack of regulation and oversight in the AI industry [which] has led to the development of chatbots that can spread misinformation and engage in hate speech" and ``the need for international regulations and agreements to ensure the safe and responsible use of AI and autonomous weapons" in the military. In addition, Mistral-7B generated impacts that are focused on the need for regulating AI in specific industries such as healthcare and law. For instance, Mistral-7B generated an impact as a result of ``the lack of regulation and oversight in the use of AI in healthcare" that can lead to ``unintended consequences and potential harm to patients" and how there is a ``need for more transparency and accountability in the development and deployment of AI systems". In the legal practice, Mistral-7B generated about ``the need for regulation and oversight to ensure the fair and ethical use of AI in the legal system" and avoid potential bias and inaccuracies in legal decisions. Other impacts generated by GPT-3 in this category also include how ``the development of AI has outpaced regulation, leading to a gap between technological advancement and governance" which may have implications for the potential misuse of AI.}
\end{enumerate}
\newpage
\begin{table*}
\subsection{Qualitative Evaluation Rubric}\label{a.2}
\footnotesize
\centering
\begin{tabular}{|l|p{6cm}|p{6cm}|}
\hline
\textbf{Criterion} & \textbf{Description} & \textbf{Evaluation Scale} \\ \hline
Validation & Evaluates whether the generated text is an impact & 
Does the generated text state or describe a negative impact of a technology? \newline
0 - The generated text is a general statement or a positive impact \newline
1 - Yes, the generated text describes/states a negative impact of a technology \\ \hline

Relevance to Stakeholders & Defined as the relevance of the negative impact to the entities and stakeholders of a technology &
How relevant is the negative impact to the entities mentioned in the functional description? \newline
1 - Irrelevant: the negative impact is irrelevant to the entities described in the functional description \newline
2 - Somewhat relevant: the negative impact could be relevant to the entities described in the functional description \newline
3 - Highly relevant: the negative impact is relevant to the entities of the technology described in the functional description \\ \hline

Relevance to Core Functionalities & Defined as the relevance of the negative impact to the functionalities of a technology &
How relevant is the negative impact to the core functionality of the technology as mentioned in the functional description? \newline
1 - Irrelevant: the negative impact is irrelevant to the core functionality described in the functional description \newline
2 - Somewhat relevant: the negative impact could be relevant to the core functionality of the technology described in the functional description \newline
3 - Highly relevant: the negative impact is relevant to the core functionality of the technology described in the functional description. \\ \hline

Coherence (Comprehensibility) & Defined in terms of comprehensibility of the generated negative impact &
Is the generated impact a complete sentence? \newline
0: No \newline
1: Yes \\ \hline

Coherence (Number of Impacts) & Defined in terms of the number of generated negative impacts &
Does the generated impact mention more than one impact in an impact statement? \newline
0: No \newline
1: Yes \\ \hline

Granularity & Defined in terms of the level of description of the generated impact &
How elaborative is the generated impact? \newline
1: Too concise (e.g., a single word) \newline
2: Could explain more (i.e., negative impact is slightly descriptive and can be elaborated on) \newline
3: Sufficient (i.e., negative impact is sufficiently descriptive) \\ \hline

Plausibility & Assesses the reasonableness that a negative impact could happen &
How reasonable is it to conclude that the generated negative impact could happen? \newline
1 - Not plausible \newline
2 - Somewhat plausible \newline
3 - Very plausible \\ \hline
\end{tabular}
\caption{Evaluation rubric of the generated text using instruction-based and fine-tuned models on coherence, relevance, granularity, and plausibility.}
\label{tab:my_label}
\end{table*}

\end{document}